\documentstyle[aclap]{article}

\title{\vspace{-0.5in}Head Automata and Bilingual Tiling:\\
Translation with Minimal Representations}

\author{Hiyan Alshawi\\
        AT\&T Research\\
        600 Mountain Avenue,
        Murray Hill, NJ 07974, USA\\
        hiyan@research.att.com}

\begin{document}

\maketitle
\vspace{-0.5in}

\begin{abstract}

We present a language model consisting of a collection
of costed bidirectional finite state automata associated
with the head words of phrases. The model is suitable for
incremental application of lexical associations in
a dynamic programming search for optimal dependency tree
derivations. We also present a model and
algorithm for machine translation involving optimal ``tiling''
of a dependency tree with entries of a costed bilingual
lexicon. Experimental results are reported comparing
methods for assigning cost functions to these models.
We conclude with a discussion of the adequacy of annotated
linguistic strings as representations for machine translation.

\end{abstract}

\section{Introduction}

Until the advent of statistical methods in the mainstream of
natural language processing, syntactic and semantic representations
were becoming progressively more complex. This trend is now
reversing itself, in part because statistical methods reduce
the burden of detailed modeling required by constraint-based
grammars,
and in part because statistical models for converting
natural language into complex syntactic or semantic
representations is not well understood at present.
At the same time, lexically centered views of language have
continued to increase in popularity. We can see this in
lexicalized grammatical theories, head-driven parsing and
generation, and statistical disambiguation based on lexical
associations.

These themes --- simple representations, statistical modeling,
and lexicalism --- form the basis for the models and algorithms
described in the bulk of this paper. The primary purpose is 
to build effective mechanisms for machine translation, the oldest
and still the most commonplace application of non-superficial
natural language processing. A secondary motivation is to test
the extent to which a non-trivial language processing task can be
carried out without complex semantic representations.

In Section~\ref{sec:headauto} we present reversible mono-lingual models
consisting of collections of simple automata associated
with the heads of phrases. These {\it head automata}
are applied by an algorithm with admissible incremental pruning
based on semantic association costs, providing
a practical solution to the problem of
combinatoric disambiguation (Church and Patil 1982).
The model is intended to combine the lexical sensitivity of
N-gram models (Jelinek et al. 1992) and the structural properties
of statistical context free grammars (Booth 1969) without the
computational overhead of statistical lexicalized tree-adjoining
grammars (Schabes 1992, Resnik 1992).
The quantitative dependency model described here grew out of
the model presented in Alshawi 1996a. An alternative model based
on transducer versions of the automata is described in Alshawi 1996b.

For translation, we
use a model for mapping dependency graphs written
by the source language head automata. This model is coded
entirely as a bilingual lexicon, with associated cost parameters.
The transfer algorithm described in Section~\ref{sec:transfer}
searches for the lowest cost `tiling' of the target dependency
graph with entries from the bilingual lexicon. Dynamic programming
is again used to make exhaustive search tractable, avoiding the
combinatoric explosion of shake-and-bake translation
(Whitelock 1992, Brew 1992).

In Section~\ref{sec:costs} we present a general framework for
associating
costs with the solutions of search processes, pointing out some
benefits of cost functions other than log likelihood, including
an error-minimization cost function for unsupervised training of
the parameters in our translation application.
Section~\ref{sec:experiment} briefly describes an English-Chinese 
translator employing the models and algorithms. We also present
experimental results comparing the performance of different
cost assignment methods.

Finally, we return to the more general discussion of
representations for machine translation and other natural
language processing tasks, arguing the case for simple
representations close to natural language itself.

\section{Head Automata Language Models}
\label{sec:headauto}

\subsection{Lexical and Dependency Parameters}

Head automata mono-lingual language models consist of
a {\it lexicon}, in which each entry is a pair $\langle w,m \rangle$ of
a word $w$ from a vocabulary $V$ and a head automaton $m$
(defined below), and a {\it parameter table} giving an assignment
of costs to events in a generative process involving the automata.

We first describe the model in terms of the familiar
paradigm of a generative statistical model, presenting the
parameters as conditional probabilities. This gives us
a stochastic version of dependency grammar (Hudson 1984).

Each derivation in the generative statistical model
produces an {\it ordered dependency tree}, that is, a tree
in which nodes dominate ordered sequences of left and right
subtrees and
in which the nodes have labels taken from the vocabulary $V$ and
the arcs have labels taken from a set $R$ of relation
symbols.
When a node with label $w$ immediately dominates a node with
label $w'$ via an arc with label $r$, we say that $w'$ is an
{\it r-dependent} of the {\it head} $w$. The interpretation
of this directed arc is that relation $r$ holds between 
particular instances of $w$ and $w'$.
(A word may have
several or no $r$-dependents for a particular relation $r$.) 
A recursive left-parent-right traversal of the nodes of an
ordered dependency tree for a derivation yields the word string
for the derivation.

A head automaton $m$ of a lexical entry $\langle w, m \rangle$
defines possible ordered local trees immediately
dominated by $w$ in derivations. Model parameters for
head automata, together with dependency parameters and
lexical parameters, give a probability distribution for
derivations.

A {\it dependency parameter}
\begin{quote}
$P(\downarrow,w'|w,r')$
\end{quote}
is the probability, given a head $w$ with a dependent arc
with label $r'$, that $w'$ is the $r'$-dependent for this arc.

A {\it lexical parameter}
\begin{quote}
$P(m,q|r,\downarrow,w)$
\end{quote}
is the probability that a local tree immediately
dominated by an $r$-dependent $w$ is derived by starting in
state $q$ of some automaton $m$ in a lexical entry $\langle w,m \rangle$.
The model also includes lexical parameters
\begin{quote}
$P(w,m,q|\rhd)$
\end{quote}
for the probability that $w$ is the head word for an entire
derivation initiated from state $q$ of automaton $m$.

\subsection{Head Automata}

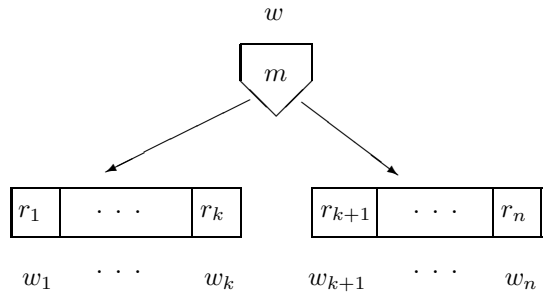
\begin{figure}
\setlength{\unitlength}{0.00083300in}
\begin{picture}(3324,1740)(664,-1654)
\put(1876,-1636){\makebox(0,0)[lb]{$w_k$}}
\put(741,-1636){\makebox(0,0)[lb]{$w_1$}}
\put(3176,-1186){\makebox(0,0)[lb]{.  .  .}}
\put(1201,-1186){\makebox(0,0)[lb]{.  .  .}}
\put(716,-1211){\makebox(0,0)[lb]{$r_1$}}
\put(2526,-1636){\makebox(0,0)[lb]{$w_{k+1}$}}
\put(3176,-1561){\makebox(0,0)[lb]{.  .  .}}
\put(1201,-1561){\makebox(0,0)[lb]{.  .  .}}
\put(2251, 14){\makebox(0,0)[lb]{$w$}}
\put(2251,-361){\makebox(0,0)[lb]{$m$}}
\put(3751,-1636){\makebox(0,0)[lb]{$w_{n}$}}
\put(3726,-1211){\makebox(0,0)[lb]{$r_n$}}
\put(2551,-1336){\framebox(1425,300){}}
\put(676,-1336){\framebox(1425,300){}}
\put(1801,-1036){\line( 0,-1){300}}
\put(976,-1036){\line( 0,-1){300}}
\put(676,-1336){\framebox(1425,300){}}
\put(2951,-1036){\line( 0,-1){300}}
\put(2601,-1211){\makebox(0,0)[lb]{$r_{k+1}$}}
\put(1851,-1211){\makebox(0,0)[lb]{$r_k$}}
\put(2161,-481){\vector(-2,-1){900}}
\put(2485,-499){\vector( 4,-3){600}}
\put(2101,-136){\line( 0,-1){225}}
\put(2101,-361){\line( 1,-1){225}}
\put(2326,-586){\line( 1, 1){225}}
\put(2551,-361){\line( 0, 1){225}}
\put(2551,-136){\line(-1, 0){450}}
\put(2101,-136){\line( 0, 1){  0}}
\put(3676,-1036){\line( 0,-1){300}}
\end{picture}
\caption{Head automaton $m$ scans left and right sequences
of relations $r_i$ for dependents $w_i$ of $w$.}
\label{fig:automaton}
\end{figure}

A head automaton is a weighted finite state machine that
writes (or accepts) a pair of sequences of relation symbols from $R$:
\begin{quote}
$(\langle r_1 \cdots r_k \rangle, \langle r_{k+1} \cdots r_n \rangle)$.
\end{quote}
These correspond to the relations between a head word and
the sequences of dependent phrases to its left and right
(see Figure~\ref{fig:automaton}).
The machine consists of a finite set $q_0, \cdots, q_s$ of states
and an
{\it action table} specifying the finite cost (non-zero
probability) actions the automaton can undergo.

There are three types of action for an automaton $m$:
left transitions, right transitions, and stop actions. These
actions, together with associated probabilistic model
parameters, are as follows.
\begin{itemize}
\item Left transition: if in state $q_{i-1}$, $m$
can write a symbol $r$ onto the right end of the
current left sequence and enter state $q_i$ with
probability $P(\leftarrow,q_i,r|q_{i-1},m)$.
\item Right transition: if in state $q_{i-1}$, $m$
can write a symbol $r$ onto the left end of the
current right sequence and enter state $q_i$ with
probability $P(\rightarrow,q_i,r|q_{i-1},m)$.
\item Stop: if in state $q$, $m$ can stop with
probability $P(\Box|q,m)$, at which point the sequences
are considered complete.
\end{itemize}
For a consistent probabilistic model, the probabilities
of all transitions and stop actions from a state $q$ must
sum to unity.
Any state of a head automaton can be an initial state,
the probability of a particular initial state in a derivation
being specified by lexical parameters.
A derivation of a pair of symbol sequence thus corresponds
to the selection of an initial state, a sequence of zero
or more transitions (writing the symbols) and a stop action.
The probability, given an initial state $q$, that automaton
$m$ will a generate a pair of sequences, i.e.
\begin{quote}
$P(\langle r_1 \cdots r_k \rangle, \langle r_{k+1} \cdots r_n \rangle | m, q)$
\end{quote}
is the product of the probabilities of the actions taken
to generate the sequences. The case of zero transitions
will yield empty sequences, corresponding to a leaf node
of the dependency tree.

From a linguistic perspective, head automata allow for a
compact, graded, notion of lexical subcategorization
(Gazdar et al. 1985) and
the linear order of a head and its dependent phrases.
Lexical parameters can control the saturation of a lexical
item (for example a verb that is both transitive and
intransitive) by starting the
same automaton in different states.
Head automata can also be used to code a grammar in
which states of an automaton for word $w$ corresponds
to X-bar levels (Jackendoff 1977) for phrases headed by $w$.

Head automata are formally more powerful than finite state
automata that accept regular languages in the following
sense. Each head automaton defines a formal language with
alphabet $R$ whose strings are the concatenation of the
left and right sequence pairs written by
the automaton. The class of languages defined in this way
clearly includes all regular languages, since strings of
a regular language can be generated, for example, by a head
automaton that only writes a left sequence.
Head automata can also accept some non-regular languages
requiring coordination of the left and right sequences,
for example the language $a^nb^n$ (requiring two states),
and the language of palindromes over a finite alphabet.

\subsection{Derivation Probability}

Let the probability of generating an ordered dependency
subtree $D$ headed by an $r$-dependent word $w$ be
$P(D|w,r)$.
The recursive process of generating this subtree proceeds
as follows: 
\begin{enumerate}
\item Select an initial state $q$ of an automaton $m$
for $w$ with lexical probability $P(m,q|r,\downarrow,w)$.
\item Run the automaton $m_0$ with initial state $q$ to
generate a pair of relation sequences with probability
$P(\langle r_1 \cdots r_k \rangle,\langle r_{k+1} \cdots r_n \rangle|m,q)$.
\item For each relation $r_i$ in these sequences,
select a dependent word $w_i$ with dependency
probability $P(\downarrow,w_i|w,r_i)$.
\item For each dependent $w_i$, recursively generate
a subtree with probability $P(D_i|w_i,r_i)$.
\end{enumerate}

We can now express the probability $P(D_0)$ for an entire
ordered dependency tree derivation $D_0$ headed by a
word $w_0$ as
\begin{quote}
$P(D_0) = \\
\hspace*{10mm} P(w_0,m_0,q_0|\rhd) \\
\hspace*{10mm} P(\langle r_1 \cdots r_k \rangle,\langle r_{k+1} \cdots r_n \rangle|m_0,q_0)\\
\hspace*{10mm} \prod_{1 \leq i \leq n}
                 P(\downarrow,w_i|w_0,r_i) P(D_i|w_i,r_i).$
\end{quote}
In the translation application we search for the
highest probability derivation (or more generally, the
N-highest probability derivations). For other purposes, the
probability of strings may be of more interest. The
probability of a string according to the model is the
sum of the probabilities of derivations of ordered
dependency trees yielding the string.

In practice, the number of parameters in a head automaton
language model is dominated by the dependency parameters,
that is, $O(|V|^2|R|)$ parameters. This puts the size of
the model somewhere in between 2-gram and 3-gram model.
The similarly motivated link grammar model (Lafferty,
Sleator and Temperley 1992) has $O(|V|^3)$ parameters.
Unlike simple N-gram models, head automata models
yield an interesting distribution of sentence lengths.
For example, the average sentence length for Monte-Carlo
generation with our probabilistic head automata model for
ATIS was 10.6 words (the average was 9.7 words for the
corpus it was trained on).

\section{Analysis and Generation}

\subsection{Analysis}

Head automaton models admit efficient lexically driven analysis
(parsing) algorithms in which partial analyses are costed
incrementally as they are constructed. Put in terms of the
traditional parsing issues in natural language understanding,
``semantic" associations coded as dependency parameters
are applied at each parsing step allowing semantically
suboptimal analyses to be eliminated,
so the analysis with the best semantic score can be identified
without scoring an exponential number of syntactic parses.
Since the model is lexical, linguistic constructions headed
by lexical items not present in the input are not involved
in the search the way they are with typical top-down or
predictive parsing strategies.

We will sketch an algorithm for finding the lowest cost
ordered dependency tree derivation for an input string
in polynomial time in the length of the string.
In our experimental system we use a more general version of
the algorithm to allow input in the form of word lattices.

The algorithm is a bottom-up tabular parser
(Younger 1967, Early 1970) in which constituents are
constructed ``head-outwards'' (Kay 1989, Sata and Stock 1989).
Since we are analyzing bottom-up with generative model
automata, the  algorithm `runs' the automata backwards.
Edges in the parsing lattice (or ``chart'')
are tuples representing partial or complete phrases
headed by a word $w$ from position $i$ to position $j$ in
the string:
\begin{quote}
$\langle w,t,i,j,m,q,c \rangle$.
\end{quote}
Here $m$ is the head automaton for $w$ in this derivation;
the automaton is in state $q$; $t$ is the dependency tree
constructed so far, and $c$ is the cost of the
partial derivation. 
We will use the notation
$C(x|y)$ for the cost of a model event with probability
$P(x|y)$; the assignment of costs to events is discussed
in Section~\ref{sec:costs}.

\noindent
{\it Initialization: }
For each word $w$ in the input between positions $i$ and $j$,
the lattice is initialized with phrases
\begin{quote}
$\langle w,\{\},i,j,m,q_f,c_f \rangle$
\end{quote}
for any lexical entry $\langle w,m \rangle$ and any final state $q_f$ of
the automaton $m$ in the entry. A final state is one for
which the stop action cost $c_f = C(\Box|q_f,m)$ is finite.

\noindent
{\it Transitions: }
Phrases are combined bottom-up to form progressively larger
phrases. There are two types of combination corresponding to
left and right transitions of the automaton for the word
acting as the head in the combination. We will specify left
combination; right combination is the mirror image of left
combination. If the lattice contains two phrases
abutting at position $k$ in the string:
\begin{quote}
$\langle w_1,t_1,i,k,m_1,q_1,c_1 \rangle$\\
$\langle w_2,t_2,k,j,m_2,q_2,c_2 \rangle$,
\end{quote}
and the parameter table contains the following finite costs
parameters (a left $r$-transition of $m_2$, a lexical parameter
for $w_1$, and an $r$-dependency parameter):
\begin{quote}
$c_3=C(\leftarrow,q_2,r|q'_2,m_2)$\\
$c_4=C(m_1,q_1|r,\downarrow,w_1)$\\
$c_5=C(\downarrow,w_1|w_2,r)$,
\end{quote}
then build a new phrase headed by $w_2$ with a tree $t_2'$ formed
by adding $t_1$ to $t_2$ as an $r$-dependent of $w_2$:
\begin{quote}
$\langle w_2,t_2',i,j,m_2,q_2',c_1+c_2+c_3+c_4+c_5 \rangle$.
\end{quote}
When no more combinations are possible, for each phrase spanning
the entire input we add the appropriate start of derivation
cost to these phrases and select the one with the lowest total
cost.

\noindent
{\it Pruning: }
The dynamic programming condition for pruning suboptimal
partial analyses is as follows. Whenever there are two phrases
\begin{quote}
$p = \langle w,t,i,j,m,q,c \rangle$\\
$p' = \langle w,t',i,j,m,q,c' \rangle$,
\end{quote}
and $c'$ is greater than $c$, then we can remove $p'$ because
for any derivation involving $p'$ that spans the entire string,
there will be a lower cost derivation involving $p$.
This pruning condition is effective at curbing a combinatorial
explosion arising from, for example, prepositional phrase
attachment ambiguities (coded in the alternative trees $t$ and
$t'$).

The worst case asymptotic time complexity of the analysis algorithm 
is $O(\mbox{min}(n^2,|V|^2) n^3)$, where $n$ is the length of an input
string and $|V|$ is the size of the vocabulary.
This limit can be derived in a similar
way to cubic time tabular recognition algorithms for context free
grammars (Younger 1967) with the grammar related term being
replaced by the term $\mbox{min}(n^2,|V|^2)$ since the words of the input
sentence also act as categories in the head automata model.
In this context ``recognition'' refers to checking that the input
string can be generated from the grammar. Note that our algorithm
is for analysis (in the sense of finding the best derivation) which,
in general, is a higher time complexity problem than recognition.

\subsection{Generation}

By generation here we mean determining the lowest cost linear surface
ordering for the dependents of each word in an {\it unordered} dependency
structure resulting from the transfer mapping
described in Section~\ref{sec:transfer}.
In general, the output of transfer
is a dependency graph and the task of the generator involves a
search for a backbone dependency tree for the graph, if necessary
by adding dependency edges to join up unconnected components of
the graph. 
For each graph component, the main steps of the search process,
described non-deterministically, are
\begin{enumerate}
\item Select a node with word label $w$ having a finite start of
derivation cost $C(w,m,q|\rhd)$.
\item Execute a path through the head automaton $m$ starting at
state $q$ and ending at state $q'$ with a finite stop action cost
$C(\Box|q',m)$. When making a transition with relation $r_i$ in
the path, select a
graph edge with label $r_i$ from $w$ to some previously unvisited
node $w_i$ with finite dependency cost $C(\downarrow,w_i|w,r_i)$.
Include the cost of the transition
(e.g. $C(\rightarrow,q_i,r_i|q_{i-1},m)$)
in the running total for this derivation.
\item For each dependent node $w_i$, select a lexical entry with
cost $C(m_i,q_i|r_i,\downarrow,w_i)$, and recursively apply the machine $m_i$
from state $q_i$ as in step 2.
\item Perform a left-parent-right traversal of the nodes of the
resulting dependency tree, yielding a target string.
\end{enumerate}
The target string resulting from the lowest cost tree that
includes all nodes in the graph is selected as the translation
target string. The independence assumptions implicit
in head automata models mean that we can select lowest cost
orderings of local dependency trees, below a given relation $r$,
independently in the search for the lowest cost derivation.

When the generator is used as part of the translation system,
the dependency parameter costs are not, in fact, applied by
the generator. Instead, because these parameters are independent
of surface order, they are applied earlier by the transfer
component, influencing the choice of structure passed to the
generator.

\section{Transfer Maps}
\label{sec:transfer}

\subsection{Transfer Model Bilingual Lexicon}

The transfer model defines possible mappings, with associated costs,
of dependency trees with source-language word node labels into
ones with target-language word labels.
Unlike the head automata monolingual models, the transfer
model operates with unordered dependency trees, that is, it treats
the dependents of a word as an unordered bag. The model is general
enough to cover the common translation problems discussed in the
literature (e.g. Lindop and Tsujii 1991 and Dorr 1994) including
many-to-many word mapping, argument switching, and head switching.

A transfer model consists of a bilingual lexicon and a transfer
parameter table.
The model uses {\it dependency tree fragments},
which are the same as unordered dependency trees except that some
nodes may not have word labels.
In the {\it bilingual lexicon}, an entry for a source word $w_i$
(see top portion of Figure~\ref{fig:tile}) has the form
\begin{quote}
$\langle w_i,H_i,n_i,G_i,f_i \rangle$
\end{quote}
where $H_i$ is a source language tree fragment,
$n_i$ (the {\it primary node}) is a distinguished node of $H_i$ with label
$w_i$, $G_i$ is a target tree fragment, and $f_i$ is a {\it mapping function},
i.e. a (possibly partial) function from the nodes of $H_i$ to the nodes
of $G_i$.

The {\it transfer parameter table} specifies costs for the application
of transfer entries. In a context-independent model, each entry has
a single cost parameter.
In context-dependent transfer models, the cost function takes
into account the identities of the labels of the arcs and nodes
dominating $w_i$ in the source graph.
(Context dependence is discussed further in Section~\ref{sec:costs}.)
The set of transfer parameters may also include costs
for the {\it null transfer entries} for $w_i$, for use in derivations
in which $w_i$ is translated by the entry for another word $v$.
For example, the entry for $v$ might be for translating an
idiom involving $w_i$ as a modifier.

Each entry in the bilingual lexicon specifies a way of mapping part
of a dependency tree, specifically that part ``matching'' (as explained
below) the source fragment of the entry, into part of a target
graph, as indicated by the target fragment.
Entry mapping functions specify how the set of target fragments
for deriving a translation are to be combined:
whenever an entry is applied, a global node-mapping function 
is extended to include the entry mapping function.

\subsection{Matching, Tiling, and Derivation}

Transfer mapping takes a source dependency tree $S$ from analysis
and produces a minimum cost derivation of a target graph $T$ and
a (possibly partial) function $f$ from source nodes to target nodes.
In fact, the transfer model is applicable to certain types of source
dependency graphs that are more general than trees, although the version
of the head automata model described here only produces trees.

We will say that a tree fragment $H$ {\it matches} an unordered
dependency tree
$S$ if there is a function $g$ (a {\it matching function}) from the
nodes of $H$ to the nodes of $S$ such that
\begin{itemize}
\item $g$ is a total one-one function;
\item if a node $n$ of $H$ has a label, and that label is word $w$,
then the word label for $g(n)$ is also $w$;
\item for every arc in $H$ with label $r$ from node $n_1$ to node
$n_2$, there is an arc with label $r$ from $g(n_1)$ to $g(n_2)$.   
\end{itemize}
Unlike first order unification, this definition of matching
is not commutative and is not deterministic in that there may be
multiple matching functions for applying a bilingual entry
to an input source tree.
A particular match of an entry against a dependency tree can be
represented by the matching function $g$, a set of arcs $A$ in
$S$, and the (possibly context dependent) cost $c$ of applying
the entry.

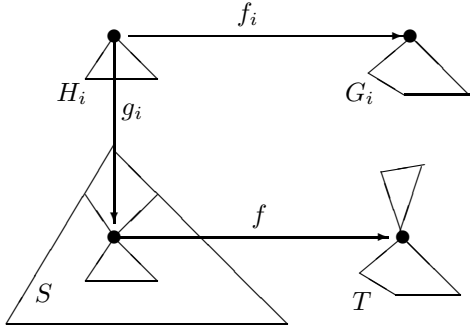
\begin{figure}
\setlength{\unitlength}{0.0005in}
\begin{picture}(4839,3270)(964,-3148)
\put(4741,-526){\line( 5,-3){375}}
\put(5116,-751){\line( 1, 0){675}}
\put(5791,-751){\line(-1, 1){600}}
\put(5191,-151){\line(-6,-5){450}}
\put(2101,-1261){\line(-3,-5){1125}}
\put(976,-3136){\line( 1, 0){2925}}
\put(3901,-3136){\line(-1, 1){1800}}
\put(5099,-2224){\line(-1, 3){225}}
\put(4874,-1549){\line( 6, 1){450}}
\put(2101,-211){\vector( 0,-1){1875}}
\put(2251,-136){\line( 1, 0){2400}}
\put(4651,-136){\vector( 1, 0){450}}
\put(2551,-586){\line(-1, 1){450}}
\put(4651,-2611){\line( 5,-3){375}}
\put(5026,-2836){\line( 1, 0){675}}
\put(5701,-2836){\line(-1, 1){600}}
\put(5101,-2236){\line(-6,-5){450}}
\put(4501,-811){\makebox(0,0)[lb]{$G_i$}}
\put(1501,-811){\makebox(0,0)[lb]{$H_i$}}
\put(4576,-2986){\makebox(0,0)[lb]{$T$}}
\put(1276,-2911){\makebox(0,0)[lb]{$S$}}
\put(2176,-961){\makebox(0,0)[lb]{$g_i$}}
\put(3376, 14){\makebox(0,0)[lb]{$f_i$}}
\put(3526,-2086){\makebox(0,0)[lb]{$f$}}
\put(1801,-586){\line( 1, 0){750}}
\put(5101,-2236){\circle*{150}}
\put(5176,-136){\circle*{150}}
\put(2101,-136){\circle*{150}}
\put(2101,-2236){\circle*{150}}
\put(2101,-2236){\line(-2,-3){300}}
\put(2101,-136){\line(-2,-3){300}}
\put(5296,-1502){\line(-1,-3){225}}
\put(2101,-2236){\line( 1, 0){2400}}
\put(4501,-2236){\vector( 1, 0){450}}
\put(2101,-2236){\line( 1, 1){450}}
\put(2101,-2236){\line(-2, 3){300}}
\put(2551,-2686){\line(-1, 1){450}}
\put(1801,-2686){\line( 1, 0){750}}
\end{picture}
\caption{Transfer matching and mapping functions}
\label{fig:tile}
\end{figure}

A {\it tiling} of a source graph with respect to a transfer
model is a set of entry matches
\begin{quote}
$\{\langle E_1, g_1, A_1, c_1 \rangle, \cdots, \langle E_k, g_k, A_k, c_k \rangle\}$
\end{quote}
which is such that
\begin{itemize}
\item $k$ is the number of nodes in the source tree $S$.
\item Each $E_i$, $1 \leq i \leq k$, is a bilingual entry
$\langle w_i,H_i,n_i,G_i,f_i \rangle$
matching $S$ with function $g_i$ (see Figure~\ref{fig:tile})
and arcs $A_i$.
\item For primary nodes $n_i$ and $n_j$ of two distinct entries
$E_i$ and $E_j$, $g_i(n_i)$ and $g_j(n_j)$ are distinct.
\item The sets of edges $A_i$ form a partition of the edges
of $S$.
\item The images $g_i(L_i)$ form a partition of the nodes of $S$,
where $L_i$ is the set of {\it labeled} source nodes in the source
fragment $H_i$ of $E_i$.
\item $c_i$ is the cost of the match specified by the
parameter table.
\end{itemize}

A tiling of $S$ yields a costed derivation of a target
dependency graph $T$ as follows:
\begin{itemize}
\item The cost of the derivation is the sum of the costs $c_i$
for each match in the tiling.
\item The nodes and arcs of $T$ are composed of the nodes and
arcs of the target fragments $G_i$ for the entries $E_i$.
\item Let $f_i$ and $f_j$ be the mapping functions for entries $E_i$
and $E_j$.
For any node $n$ of $S$ for which target nodes $f_i(g_i^{-1}(n))$
and $f_j(g_i^{-1}(n))$ are defined, these two nodes are identified as a
single node $f(n)$ in $T$.
\end{itemize}

The merging of target fragment nodes in the last condition
has the effect of joining the target fragments in a consistent
fashion.
The node mapping function $f$ for the entire tree thus has a different
role from the alignment function in the IBM statistical translation model
(Brown et al. 1990, 1993); the role of the latter includes
the linear ordering of words in the target string.
In our approach, target word order is handled exclusively by the
target monolingual model.

\subsection{Transfer Algorithm}

The main transfer search is preceded by a bilingual lexicon matching
phase. This leads to greater efficiency as it avoids
repeating matching operations during the search phase, and it
allows a static analysis of the matching entries and source tree
to identify subtrees for which the search phase can safely prune
out suboptimal partial translations.

\paragraph*{Transfer Configurations}

In order to apply target language model relation costs incrementally,
we need to distinguish between complete and incomplete arcs:
an arc is complete if both its nodes have labels, otherwise it is
incomplete.
The output of the lexicon matching phrase, and the partial derivations
manipulated by the search phase are both in the form of {\it transfer
configurations}
\begin{quote}
$\langle S, R, T, P, f, c, I \rangle$ 
\end{quote}
where $S$ is the set of source nodes and arcs consumed so far in
the derivation, $R$ the remaining source nodes and arcs, $f$ the
mapping function built so far, $T$ the set of nodes and complete
arcs of the target graph, $P$ the set of incomplete target arcs,
$c$ the partial derivation cost, and $I$ a set
of source nodes for which entries have yet to be applied.

\paragraph*{Lexical matching phase}

The algorithm for lexical matching has a similar control structure to
standard unification algorithms, except that it can result in multiple
matches. We omit the details. 
The lexicon matching phase returns, for each source node $i$, a set of
{\it runtime entries}. There is one runtime entry for each
successful match and possibly a null entry for the node if the word
label for $i$ is included in successful matches for other entries.
Runtime entries are transfer configurations of the form
\begin{quote}
$\langle H_i, \phi, G_i, P_i, f_i, c_i, \{i\} \rangle$ 
\end{quote}
in which $H_i$ is the source fragment for the entry with each node
replaced by its image under the applicable matching function;
$G_i$ the target fragment for the entry, except for the incomplete
arcs $P_i$ of this fragment; $f_i$ the composition of mapping function
for the entry with the inverse of the matching function;
$c_i$ the cost of applying the entry in the context of its match with
the source graph plus the cost in the target model of the arcs in
$G_i$.

\paragraph*{Transfer Search}

Before the transfer search proper, the resulting runtime entries
together with the source graph are analyzed to determine
{\it decomposition nodes}. A decomposition node $n$ is a source tree
node for which it is safe to prune suboptimal translations of the
subtree dominated by $n$. Specifically, it is checked that $n$ is
the root node of all source fragments $H_n$ of runtime entries in which
both $n$ {\it and} its node label are included, and that $f_n(n)$ is
not dominated by (i.e. not reachable via directed arcs from)
another node in the target graph $G_n$ of such entries.

Transfer search maintains a set $M$ of active runtime entries. Initially,
this is the set of runtime entries resulting from the lexicon matching
phase. Overall search control is as follows:
\begin{enumerate}
\item Determine the set of decomposition nodes.
\item Sort the decomposition nodes into a list $D$ such that if $n_1$
dominates $n_2$ in $S$ then $n_2$ precedes $n_1$ in $D$.
\item If $D$ is empty, apply the subtree transfer search (given below)
to $S$, return the lowest cost solution, and stop.
\item Remove the first decomposition node $n$ from $D$ and apply the
subtree transfer search to the subtree $S'$ dominated by $n$, to yield
solutions\\
$\langle S',\phi,T',\phi,f',c',\phi \rangle$.
\item Partition these solutions into subsets with the same word label for
the node $f'(n)$, and select the solution with lowest cost
$c'$ from each subset.
\item Remove from $M$ the set of runtime entries for nodes in $S'$.
\item For each selected subtree solution, add to $M$ a new runtime
entry $\langle S', \phi, T', f', c', \{n\} \rangle$.
\item Repeat from step 3.
\end{enumerate}

The subtree transfer search maintains a queue $Q$ of configurations
corresponding to partial derivations for translating the subtree.
Control follows a standard non-deterministic search paradigm:
\begin{enumerate}
\item Initialize $Q$ to contain a single configuration\\
$\langle \phi, R_0, \phi, \phi, \phi, 0, I_0 \rangle$
with the input subtree $R_0$ and the set of nodes $I_0$ in $R_0$.
\item If $Q$ is empty, return the lowest cost solution found and stop.
\item Remove a configuration $\langle S, R, T, P, f, c, I \rangle$ from the queue.
\item If $R$ is empty, add the configuration to the set of subtree
solutions.
\item Select a node $i$ from $I$.
\item For each runtime entry $\langle H_i, \phi, G_i, P_i, f_i, c_i, \{i\} \rangle$
for $i$, if $H_i$ is a subgraph of $R$, add to $Q$ a configuration
$\langle S \cup H_i, R-H_i, T \cup G_i \cup G', P \cup P_i-G', f \cup f_i, c+c_i+c_{G'}, I-\{i\} \rangle$,
where $G'$ is the set of newly completed arcs (those in $P \cup P_i$ with
both node labels in $T \cup G_i \cup P \cup P_i$) and $c_{G'}$ is the cost of the arcs $G'$
in the target language model.
\item For any source node $n$ for which $f(n)$ and $f_i(n)$ are both
defined, merge these two target nodes.
\item Repeat from step 2.
\end{enumerate}
Keeping the arcs $P$ separate in the configuration allows efficient
incremental application of target dependency costs $c_{G'}$ during
the search, so these costs are taken into account in the pruning
step of the overall search control.
This way we can keep the benefits of monolingual/bilingual modularity
(Isabelle and Macklovitch 1986) without the computational overhead
of transfer-and-filter (Alshawi et al. 1992).

It is possible to apply the subtree search directly to the
whole graph starting with the initial runtime entries from lexical
matching. However, this would result in an exponential search, specifically
a search tree with a branching factor of the order of the number of
matching entries per input word. Fortunately, long sentences typically
have several decomposition nodes, such as the heads of noun
phrases, so the search as described is factored into manageable components.

\section{Cost Functions}
\label{sec:costs}

\subsection{Costed Search Processes}

The head automata model and transfer model were
originally conceived as probabilistic models. In order
to take advantage of more of the information available
in our training data, we experimented with cost functions
that make use of incorrect translations as negative
examples and also to treat the correctness of a
translation hypothesis as a matter of degree.

To experiment with different models, we implemented a
general mechanism for associating costs to solutions of
a search process.
Here, a search process is conceptualized as a
non-deterministic computation that takes a single input
string, undergoes a sequence of state transitions in
a non-deterministic fashion, then outputs a solution
string.
Process states are distinct from, but may include,
head automaton states.

A cost function for a search process is a real valued
function defined on a pair of equivalence classes of
process states.
The first element of the pair, a {\it context} $c$, is
an equivalence class of states before transitions. The
second element, an {\it event} $e$, is an equivalence class
of states after transitions. (The equivalence relations
for contexts and events may be different.)
We refer to an event-context pair
as a {\it choice}, for which we use the notation
\begin{quote}
$(e | c)$
\end{quote}
borrowed from the special case of conditional
probabilities. The cost of a derivation of a solution
by the process is taken to be the sum of costs of
choices involved in the derivation.

We represent events and contexts by finite
sequences of symbols (typically words or relation
symbols in the translation application). We write
\begin{quote}
$C(a_1 \cdots a_n | b_1 \cdots b_k)$
\end{quote}
for the cost of the event represented
by $\langle a_1 \cdots a_n \rangle$ in the context represented
by$\langle b_1 \cdots b_k \rangle$.

``Backed off'' costs can be computed
by averaging over larger equivalence classes (represented
by shorter sequences in which positions are eliminated
systematically). A similar smoothing technique has
been applied to the specific case of prepositional phrase
attachment by Collins and Brooks (1995).
We have used backed off costs in the translation application
for the various cost functions described below. Although
this resulted in some improvement in testing, so far
the improvement has not been statistically significant.

\subsection{Model Cost Functions}

Taken together, the events, contexts, and cost function constitute
a {\em process cost model}, or simply a {\it model}.
The cost function specifies the {\it model parameters}; the other
components are the {\it model structure}.

We have experimented with a number of model types,
including the following.

\noindent
{\it Probabilistic model: }
In this model we assume a
probability distribution on the possible events for a context,
that is,
\begin{quote}
$\sum_{e} P(e | c) = 1$.
\end{quote}
The cost parameters of the model are defined as:
\begin{quote}
$C(e | c) = -\ln(P(e | c))$.
\end{quote}
Given a set of solutions from executions of a process,
let $n^+(e|c)$ be the number of times
choice $(e|c)$ was taken leading to acceptable solutions
(e.g. correct translations) and $n^+(c)$ be the number of times
context $c$ was encountered for these solutions.
We can then estimate the probabilistic model costs with 
\begin{quote}
$C(e | c) \approx \ln(n^+(c)) - \ln(n^+(e | c))$.
\end{quote}

\noindent
{\it Discriminative model: }
The costs in this model
are likelihood ratios comparing positive and negative solutions,
for example correct and incorrect translations. (See Dunning 1993
on the application of likelihood ratios in computational
linguistics.)
Let $n^-(e|c)$ be the count for choice $(e|c)$
leading to negative solutions. The cost function for the
discriminative model is estimated as
\begin{quote}
$C(e | c) \approx \ln(n^-(e | c)) - \ln(n^+(e | c))$.
\end{quote}

\noindent
{\it Mean distance model: }
In the mean distance model,
we make use of some measure of goodness of a
solution $t_s$ for some input $s$ by comparing it against an
ideal solution $\hat{t}_s$ for $s$ with a distance metric $h$:
\begin{quote}
$h(t_s,\hat{t}_s) \mapsto d$
\end{quote}
in which $d$ is a non-negative real number.
A parameter for choice $(e|c)$ in the distance model
\begin{quote}
$C(e|c) = E_h(e|c)$
\end{quote}
is the mean value of $h(t_s,\hat{t}_s)$ for solutions $t_s$ produced
by derivations including the choice $(e|c)$.

\noindent
{\it Normalized distance model: }
The mean distance model does not use the constraint that
a particular choice faced by a process is always a choice between
events with the same context. It is also somewhat sensitive
to peculiarities of the distance function $h$. With the same
assumptions we made  for the mean distance model, let
\begin{quote}
$E_h(c)$
\end{quote}
be the average of $h(t_s,\hat{t}_s)$ for solutions derived from
sequences of choices including the context $c$. The cost
parameter for $(e|c)$ in the normalized distance model is
\begin{quote}
$C(e | c) = \frac{E_h(e|c)}{E_h(c)}$,
\end{quote}
that is, the ratio of the expected distance for derivations
involving the choice and the expected distance for
all derivations involving the context for that choice.

\paragraph*{Reflexive Training}

If we have a manually translated corpus, we can apply the
mean and normalized distance models to translation by taking
the ideal solution $\hat{t}_s$ for translating a source string $s$
to be the manual translation for $s$.
In the absence of good metrics for comparing translations,
we employ a heuristic string distance metric to compare
word selection and word order in $t_s$ and $\hat{t}_s$.

In order to train the model parameters without a manually
translated corpus, we use a ``reflexive'' training
method (similar in spirit to the ``wake-sleep'' algorithm,
Hinton et al. 1995). In this method, our search process translates
a source sentence $s$ to $t_s$ in the target language and then
translates $t_s$ back to a source language sentence $s'$.
The original sentence $s$ can then act as the ideal solution
of the overall process.
For this training method to be effective,
we need a reasonably good initial model, i.e. one for which
the distance $h(s,s')$ is inversely correlated with the
probability that $t_s$ is a good translation of $s$.

\section{Experimental System}
\label{sec:experiment}

We have built an experimental translation system using the
monolingual and translation models described in this paper.
The system translates sentences in the ATIS
domain (Hirschman et al. 1993) between English and
Mandarin Chinese.
The translator is in fact a
subsystem of a speech translation prototype, though the
experiments we describe here are for transcribed spoken
utterances. (We informally refer to the transcribed utterances
as sentences.)
The average time taken for translation of sentences
(of unrestricted length) from the ATIS corpus was around 1.7
seconds with approximately 0.4 seconds being taken by the
analysis algorithm and 0.7 seconds by the transfer algorithm.

English and Chinese lexicons of around 1200 and
1000 words respectively were constructed. Altogether, the
entries in these lexicons made reference to around 200
structurally distinct head automata. The transfer lexicon
contained around 3500 paired graph fragments, most of which
were used in both transfer directions.
With this model structure, we tried
a number of methods for assigning cost functions.
The nature of the training methods and their corresponding
cost functions meant that different amounts of training data
could be used, as discussed further below.

The methods make use of a supervised training set and
an unsupervised training set, both sets being
chosen at random from the 20,000 or so ATIS sentences available to
us. The supervised training set comprised around 1950 sentences.
A subcollection of 1150 of these sentences were translated
by the system, and the resulting translations manually
classified as `good' (800 translations) or `bad' (350
translations).
The remaining 800 supervised training set sentences were
hand-tagged for prepositional attachment points.
(Prepositional phrase attachment is a major cause of ambiguity
in the ATIS corpus, and moreover can affect English-Chinese
translation, see Chen and Chen 1992.)
The attachment information was used to generate
additional negative and positive counts for dependency choices.
The unsupervised training set consisted of approximately
13,000 sentences; it was used for automatic training (as described
under `Reflexive Training' above) by translating the sentences
into Chinese and back to English.

\noindent
A. {\it Qualitative Baseline: }
In this model, all choices were assigned the same cost
except for irregular events (such as
unknown words or partial analyses) which
were all assigned a high penalty cost. This model gives an
indication of performance based solely on model structure.

\noindent
B. {\it Probabilistic: }
Counts for choices leading to
good translations for sentences of the supervised training
corpus, together with counts from the manually assigned
attachment points, were used to compute negated log probability
costs.

\noindent
C. {\it Discriminative: }
The positive counts as in
the probabilistic method, together with corresponding
negative counts from bad translations or incorrect attachment
choices, were used to compute log likelihood ratio costs.

\noindent
D. {\it Normalized Distance: }
In this fully automatic method,
normalized distance costs were computed from reflexive
translation of the sentences in the unsupervised training corpus.
The translation runs were carried out with parameters from
method A.

\noindent
E. {\it Bootstrapped Normalized Distance: }
The same as
method D except that the system used to carry out the
reflexive translation was running with parameters from
method C.

Table~\ref{tab:methods} shows the results of evaluating the performance of
these models for translating 200 unrestricted length ATIS sentences
into Chinese.
This was a previously unseen test set not included in any of the
training sets. Two measures of translation acceptability are
shown, as judged by a Chinese speaker. (In separate
experiments, we verified that the judgments of this speaker
were near the average of five Chinese speakers).
The first measure, ``meaning and grammar'', gives the percentage
of sentence translations judged to preserve meaning without the
introduction of grammatical errors. For the second measure, ``meaning 
preservation'', grammatical errors were allowed if they
did not interfere with meaning (in the sense of misleading
the hearer).
\begin{table}
\caption{Translation performance of different cost assignment methods}
\vspace*{2mm}
\begin{tabular}{|l|c|c|}\hline
Method & Meaning and  & Meaning \\
       & Grammar (\%) & Preservation (\%) \\ \hline
A & 29 & 71 \\
D & 37 & 71 \\ \hline
B & 46 & 82 \\
C & 52 & 83 \\
E & 54 & 83 \\ \hline
\end{tabular}
\label{tab:methods}
\end{table}
In the table, we have grouped together methods A and D for
which the parameters were derived without human supervision
effort, and methods B, C, and E which depended on
the same amount of human supervision effort. This means that side
by side comparison of these methods has practical relevance,
even though the methods exploited different amounts of data.
In the case of E, the supervision
effort was used only as an oracle during training, not
directly in the cost computations.

We can see from Table~\ref{tab:methods} that the choice of method affected
translation quality (meaning and grammar) more than it affected
preservation of meaning. A possible explanation is that
the model structure was adequate for most lexical choice
decisions because of the
relatively low degree of polysemy in the ATIS corpus. 
For the stricter measure, the differences were
statistically significant, according to the sign test at
the 5\% significance level, for the following comparisons:
C and E each outperformed B and D, and B and D each
outperformed A.

\section{Language Processing and Semantic Representations}

The translation system we have described employs
only simple representations of sentences and phrases.
Apart from the words themselves, the only
symbols used are the dependency relations $R$. In our
experimental system, these relation symbols are themselves
natural language words, although this is not a necessary
property of our models.
Information coded explicitly in sentence representations
by word senses and feature constraints in our previous
work (Alshawi 1992) is implicit in the models used to
derive the dependency trees and translations.
In particular, dependency parameters and context-dependent
transfer parameters give rise to an implicit, graded notion
of word sense.

For language-centered applications like translation or
summarization, for which we have a large body of examples of
the desired behavior, we can think of the task in terms of
the formal problem of modeling a relation between strings
based on examples of that relation.
By taking this viewpoint, we seem to be ignoring the intuition
that most interesting natural language processing tasks
(translation, summarization, interfaces) are semantic in nature.
It is therefore tempting to conclude that an adequate treatment
of these tasks requires the manipulation of artificial
semantic representation languages with well-understood formal
denotations. While the intuition seems reasonable, the
conclusion might be too strong in that it rules out the
possibility that natural language itself is adequate for
manipulating semantic denotations. After all, this is the
primary function of natural language.

The main justification for artificial semantic representation
languages is that they are unambiguous by design. This may not
be as critical, or useful, as it might first appear.
While it is true that
natural language is ambiguous and under-specified out of context,
this uncertainty is greatly reduced by context to the point
where further resolution (e.g. full scoping) is irrelevant to
the task, or even the intended meaning. The fact that translation
is insensitive to many ambiguities motivated the use of
unresolved quasi-logical form for transfer (Alshawi et al. 1992).

To the extent that contextual resolution is necessary, context may
be provided by the state of the language processor rather
than complex semantic representations.
Local context may include the state of
local processing components (such as our head automata) for capturing
grammatical constraints, or the identity of other words in a phrase
for capturing sense distinctions. For larger scale context,
I have argued elsewhere (Alshawi 1987) that memory activation patterns
resulting from the process of carrying out an understanding task can
act as global context without explicit representations of discourse.
Under this view,
the challenge is how to exploit context in performing
a task rather than how to map natural language phrases to
expressions of a formalism for coding meaning
independently of context or intended use.

There is now greater understanding of the formal semantics of
under-specified and ambiguous representations. In Alshawi 1995, I
provide a denotational semantics for a simple under-specified
language and argue for extending this treatment to a formal
semantics of natural language strings as expressions of an
under-specified representation. In this paradigm, ordered dependency
trees can be viewed as natural language strings annotated
so that some of the implicit relations are more explicit. A milder
form of this kind of annotation is a bracketed natural language
string.
We are not advocating an approach in which linguistic
structure is ignored (as it is in the IBM translator described
by Brown et al. 1990), but rather one in which the syntactic and
semantic structure of a string is implicit in the way it
is processed by an interpreter.

One important advantage of using representations that
are close to natural language itself is that it reduces
the degrees of freedom in specifying language and task models,
making these models easier to acquire automatically.
With these considerations in mind, we have started to
experiment with a version of the translator described here
with even simpler representations and for which the
model structure, not just the parameters, can be acquired
automatically.

\section*{Acknowledgments}

The work on cost functions and training methods was carried
out jointly with Adam Buchsbaum who also customized the
English model to ATIS and integrated the translator into
our speech translation prototype.
Jishen He constructed the Chinese ATIS language model and
bilingual lexicon and identified many problems with
early versions of the transfer component. I am also grateful
for advice and help from Don Hindle, Fernando Pereira,
Chi-Lin Shih, Richard Sproat, and Bin Wu.

\section*{References}

\newenvironment{reverseindent}%
{\begin{list}{}{\setlength{\labelsep}{0in} 
                \setlength{\labelwidth}{0in}
                \setlength{\itemindent}{-\leftmargin}}}%
{\end{list}}

\begin{reverseindent}

\item\pagebreak[3]
Alshawi, H. 1996a. ``Qualitative and Quantitative Models of Speech
Translation''. In {\it The Balancing Act: Combining
Symbolic and Statistical Approaches to Language}, edited
by P. Resnik and J. Klavans, The MIT Press, Cambridge, Massachusetts.

\item\pagebreak[3]
Alshawi, H. 1996b. ``Head Automata for Speech Translation''.
In {\it Proceedings of ICSLP 96, the Fourth International
Conference on Spoken Language Processing}, Philadelphia,
Pennsylvania.

\item\pagebreak[3]
Alshawi, H. 1995. ``Underspecified First Order Logics''.
In {\it Semantic Ambiguity and Underspecification},
edited by K. van Deemter and S. Peters, CSLI Publications,
Stanford, California.

\item\pagebreak[3]
Alshawi, H. 1992. {\it The Core Language Engine}.
MIT Press, Cambridge, Massachusetts.

\item\pagebreak[3]
Alshawi, H. 1987.
{\it Memory and Context for Language Interpretation}.
Cambridge University Press, Cambridge, England.

\item\pagebreak[3]
Alshawi, H., D. Carter, B. Gamback and M. Rayner. 1992.
``Swedish-English QLF Translation''.
In {\it The Core Language Engine}, edited by H. Alshawi,
The MIT Press, Cambridge, Massachusetts.

\item\pagebreak[3]
Booth, T. 1969. ``Probabilistic Representation of Formal Languages''.
{\it Tenth Annual IEEE Symposium on Switching and Automata Theory}.

\item\pagebreak[3]
Brew, C. 1992. ``Letting the Cat out of the Bag: Generation for
Shake-and-Bake MT''. Proceedings of COLING92, the International
Conference on Computational Linguistics, Nantes, France.

\item\pagebreak[3]
Brown, P., J. Cocke, S. Della Pietra, V. Della Pietra, F. Jelinek,
J. Lafferty, R. Mercer and P. Rossin. 1990.
``A Statistical Approach to Machine Translation''.
{\it Computational Linguistics} 16:79--85.

\item\pagebreak[3]
Brown, P.F., S.A. Della Pietra, V.J. Della Pietra,
and R.L. Mercer. 1993.
``The Mathematics of Statistical Machine Translation: Parameter
Estimation''.
{\it Computational Linguistics} 19:263--312.

\item\pagebreak[3]
Chen, K.H. and H. H. Chen. 1992. ``Attachment and 
Transfer of Prepositional Phrases with
Constraint Propagation''. {\it Computer Processing
of Chinese and Oriental Languages}, Vol. 6, No. 2,
123--142.

\item\pagebreak[3]
Church K. and R. Patil. 1982.
``Coping with Syntactic Ambiguity or How to Put the Block in the
Box on the Table''.
{\it Computational Linguistics} 8:139--149.

\item\pagebreak[3]
Collins, M. and J. Brooks. 1995.
``Prepositional Phrase Attachment through a Backed-Off Model.''
{\it Proceedings of the Third Workshop on Very Large Corpora},
Cambridge, Massachusetts, ACL, 27--38.

\item\pagebreak[3]
Dorr, B.J. 1994.
``Machine Translation Divergences: A Formal
Description and Proposed Solution''.
{\it Computational Linguistics} 20:597--634.

\item\pagebreak[3]
Dunning, T. 1993. ``Accurate Methods for Statistics
of Surprise and Coincidence.'' {\it Computational Linguistics}.
19:61--74.

\item\pagebreak[3]
Early, J. 1970.
``An Efficient Context-Free Parsing Algorithm''.
{\it Communications of the ACM} 14: 453--60.

\item\pagebreak[3]
Gazdar, G., E. Klein, G.K. Pullum, and I.A.Sag. 1985.
{\it Generalised Phrase Structure Grammar}. Blackwell, Oxford.

\item\pagebreak[3]
Hinton, G.E., P. Dayan, B.J. Frey and R.M. Neal. 1995.
``The `Wake-Sleep' Algorithm for Unsupervised Neural Networks''.
{\it Science} 268:1158--1161.

\item\pagebreak[3]
Hudson, R.A. 1984. {\it Word Grammar}. Blackwell, Oxford.

\item\pagebreak[3]
Hirschman, L., M. Bates, D. Dahl, W. Fisher, J. Garofolo,
D. Pallett, K. Hunicke-Smith, P. Price, A. Rudnicky,
and E. Tzoukermann. 1993.
``Multi-Site Data Collection and Evaluation in Spoken Language
Understanding''.
In {\it Proceedings of the Human Language Technology Workshop},
Morgan Kaufmann, San Francisco, 19--24.

\item\pagebreak[3]
Isabelle, P. and E. Macklovitch. 1986. ``Transfer and MT
Modularity'', {\it Eleventh International Conference on Computational
Linguistics}, Bonn, Germany, 115--117.

\item\pagebreak[3]
Jackendoff, R.S. 1977. {\it X-bar Syntax: A Study of Phrase Structure}.
MIT Press, Cambridge, Massachusetts.

\item\pagebreak[3]
Jelinek, F., R.L. Mercer and S. Roukos. 1992. ``Principles of Lexical
Language Modeling for Speech Recognition''.
In {\it Advances in Speech Signal Processing},
edited by S. Furui and M.M. Sondhi. Marcel Dekker, New York.

\item\pagebreak[3]
Lafferty, J., D. Sleator and D. Temperley. 1992. ``Grammatical
Trigrams: A Probabilistic Model of Link Grammar''. In {\it Proceedings
of the 1992 AAAI Fall Symposium on Probabilistic Approaches to
Natural Language}, 89-97.

\item\pagebreak[3]
Kay, M. 1989.
``Head Driven Parsing''.
In {\it Proceedings of the Workshop on Parsing Technologies},
Pittsburg, 1989.

\item\pagebreak[3]
Lindop, J. and J. Tsujii. 1991. ``Complex Transfer in MT: A Survey
of Examples''. Technical Report 91/5, Centre for Computational
Linguistics, UMIST, Manchester, UK. 

\item\pagebreak[3]
Resnik, P. 1992. ``Probabilistic Tree-Adjoining Grammar as a Framework
for Statistical Natural Language Processing''. In {\it Proceedings of
COLING-92}, Nantes, France, 418-424.

\item\pagebreak[3]
Sata, G. and O. Stock. 1989.
``Head-Driven Bidirectional Parsing''.
In {\it Proceedings of the Workshop on Parsing Technologies},
Pittsburg, 1989.

\item\pagebreak[3]
Schabes, Y. 1992. ``Stochastic Lexicalized Tree-Adjoining Grammars''.
In {\it Proceedings of COLING-92}, Nantes, France, 426-432.

\item\pagebreak[3]
Whitelock, P.J. 1992. ``Shake-and-Bake Translation''.  Proceedings of
COLING92, the International Conference on Computational Linguistics,
Nantes, France.

\item\pagebreak[3]
Younger, D. 1967. Recognition and Parsing of Context-Free Languages
in Time $n^3$. {\it Information and Control}, 10, 189--208.

\end{reverseindent}

\end{document}